\documentclass[conference]{IEEEtran}
\IEEEoverridecommandlockouts
\usepackage{cite}
\usepackage{amsmath,amssymb,amsfonts}
\usepackage{algorithmic}
\usepackage{graphicx}
\usepackage{textcomp}
\usepackage{xcolor}
\usepackage{subfig}
\usepackage{multirow}
\usepackage[para,online,flushleft]{threeparttable}

\def\BibTeX{{\rm B\kern-.05em{\sc i\kern-.025em b}\kern-.08em
    T\kern-.1667em\lower.7ex\hbox{E}\kern-.125emX}}

\begin{document}

\title{Elastic Neural Networks for Classification\\
\thanks{This work was partially funded by the Academy of Finland
project 309903 CoefNet. Authors also thank CSC--IT Center for Science for computational resources.}
% \author{Paper ID (given by CMT)} %Comment this line for camera-ready version of accepted paper
\author{Yi Zhou$^{1}$, Yue Bai$^{1}$, Shuvra S. Bhattacharyya$^{1},^{2}$ and Heikki Huttunen$^{1}$\\
$^{1}$Tampere University of Technology, Finland,  $^{2}$University of Maryland, USA}
}

\maketitle

\begin{abstract}

In this work we propose a framework for improving the performance of any deep neural network that may suffer from vanishing gradients. To address the vanishing gradient issue, we study a framework, where we insert an intermediate output branch after each layer in the computational graph and use the corresponding prediction loss for feeding the gradient to the early layers. The framework---which we name Elastic network---is tested with several well-known networks on CIFAR10 and CIFAR100 datasets, and the experimental results show that the proposed framework improves the accuracy on both shallow networks (\textit{e.g.,} MobileNet) and deep convolutional neural networks (\textit{e.g.,} DenseNet). We also identify the types of networks where the framework does not improve the performance and discuss the reasons. Finally, as a side product, the computational complexity of the resulting networks can be adjusted in an elastic manner by selecting the output branch according to current computational budget.

% Deep learning has been widely applied in various computer vision fields. Deeper and deeper convolutional neural networks are perused and addressed attentions in recent researches. However, the increasing of layers and complexities of convolution neural networks generates problems such as over-parameterized and over-fitting which indeed does harm to the network performance. 
\end{abstract}


\section{Introduction}

Deep convolutional neural networks (DCNNs) are currently the state of the art approach in visual object recognition tasks. DCNNs have been extensively applied on classification tasks since the start of \textit{ILSVRC} challenge \cite{russakovsky2015imagenet}. During the history of this competition, probably the only sustained trend has been the increase in network depth: AlexNet\cite{krizhevsky2012imagenet} with 8 layers won the first place in 2012 with an top-5 error rate of 16\%; VGG\cite{simonyan2014very} consists of 16 layers won the first place and decreased error rate to 7.3\% in 2014; In 2015, ResNet \cite{he2016deep} with 152 very deep convolutional layers and identity connections won the first place with continuously decreasing the error rate to 3.6\%. %The accuracy raised with the increasing depth of neural networks. However, the over-parameterized and the over-fitting problems have been continuously carrying on. 

Thus, the depth of neural network seems to correlate with the accuracy of the network. However, extremely deep networks (over 1,000 layers) are challenging to train and are not widely in use yet. One of the reasons for their mediocre performance is the behaviour of the gradient at the early layers of the network. More specifically, as gradient is passed down the computational graph for weight update, its magnitude either tends to decrease (vanishing gradient) or increase (exploding gradient), making the gradient update either very slow or unstable. There are a number of approaches to avoid these problems, such as using the rectified linear unit (ReLU) activation \cite{nair2010rectified} or controlling the layer activations using batch normalization \cite{ioffe2015batch}. However, the mainstream of research concentrates on how to \textit{preserve} the gradient, while less research has been done on how to \textit{feed} the gradient via direct pathways.

%In \cite{romero2014fitnets} reported that it was only possible to train Maxout networks with up to 5 layers through plain backpropagation. 
% \vspace*{\fill}
% This work was partially funded by the Academy of Finland project 309903 CoefNet. Authors also thank CSC–IT Center for Science for computational
% resources.

On the other hand, computationally lightweight neural networks are subject to increasing interest, as they are widely used in industrial and real-time applications such as self-driving cars. What is still missing, however, is the flexibility to adjust to changing computational demands in a flexible manner. 

%question: about two main approaches, describe in detail. Can be extended

To address both issues---lightweight and very deep neural networks---we propose a flexible architecture called Elastic Net, which can be easily applied on any existing convolutional neural networks. The key idea is to add auxiliary outputs to the intermediate layers of the network graph, and train the network against the joint loss over all layers. We will illustrate that this simple idea of adding intermediate outputs, enables the Elastic Net to seamlessly switch between different levels of computational complexities while simultaneously achieving improved accuracy (compared to the backbone network without intermediate outputs) when a high computational budget is available.

%With the integrating of low-level features (which we usually get from shallow layers) together with high level features, JB structured neural networks take extra regularizers and show more generalization ability. The architecture of the proposed JB Net and the methodology will be further discussed in Section III.

We study the Elastic Nets for classification problem and test our approach on two classical datasets CIFAR10 and CIFAR100. Elastic Nets can be constructed on top of both shallow (\textit{e.g.,} VGG, MobileNet \cite{howard2017mobilenets}) and very deep (\textit{e.g.,} DenseNet, InceptionV3) architectures. Our proposed Elastic Nets show better performance on most of the networks above. Details of the experiment design and networks training will be explained in Section IV. 
%introduce data and networks.

%In this paper, we propose a general architecture for neural network models. The flexibility of the JB neural network architecture means that it can be easily applied both to very deep neural networks and lightweight convolutional neural networks. 
%JB Net don't change the main network architecture and can be easily applied to other neural models by adding auxiliary classifiers to achieve higher accuracy. 
%We attaching classifiers to the intermediate layers of different models. 
%By adding intermediate classifiers, model can combine low-level features (usually we get from shallow layer) and high level features. 
%Gradient vanishing can be mitigated by joint loss backpropagation and attaching intermediate classifiers can be considered one method of regularization. %(it's possible to explain the result from this perspective)%

Although attaching intermediate outputs to the network graph has been studied earlier, \cite{guo2016shallow,teerapittayanon2016branchynet,bai2018elastic}, we propose a general \textit{framework} that applies to \textit{any network} instead of a new network structure. Moreover, the intermediate outputs are added in systematic manner instead of hand-tuning the network topology.
%The key contributions in our paper are summarized as follows.

% \begin{itemize}
  
%   \item Our proposed approach introduces more intermediate outputs than any previous works. There were at the most 5 intermediate outputs in all previous works. On the contrary, in our paper, we insert the number of intermediate outputs even more (e.g. 11 intermediate outputs in Inception V3).
% %question: how many intermediate outputs in Inception V3?
% %For Inception V3, 11 intermediate classifiers are added instead of 2 classifiers in original Inception V3 model.
  
%   \item We show that JB structured networks mitigate the over-parametrized problem in very deep networks. We also introduce our JB Net with lightweight network (e.g. MobileNet) and achieves better performance. 
% %For MobileNet, we add 16 classifiers between the fist convolutional layer and the last FC layer. The accuracy improvement is most obvious in lightweight model than normal one.
  
%   \item We show that JB structured networks can generalize better and improve the accuracy for all studied DCNNs, except ResNet50 and VGG.
  
% \end{itemize}

\begin{figure*}[!h]
 \begin{center}
 \includegraphics[width=0.9\linewidth]{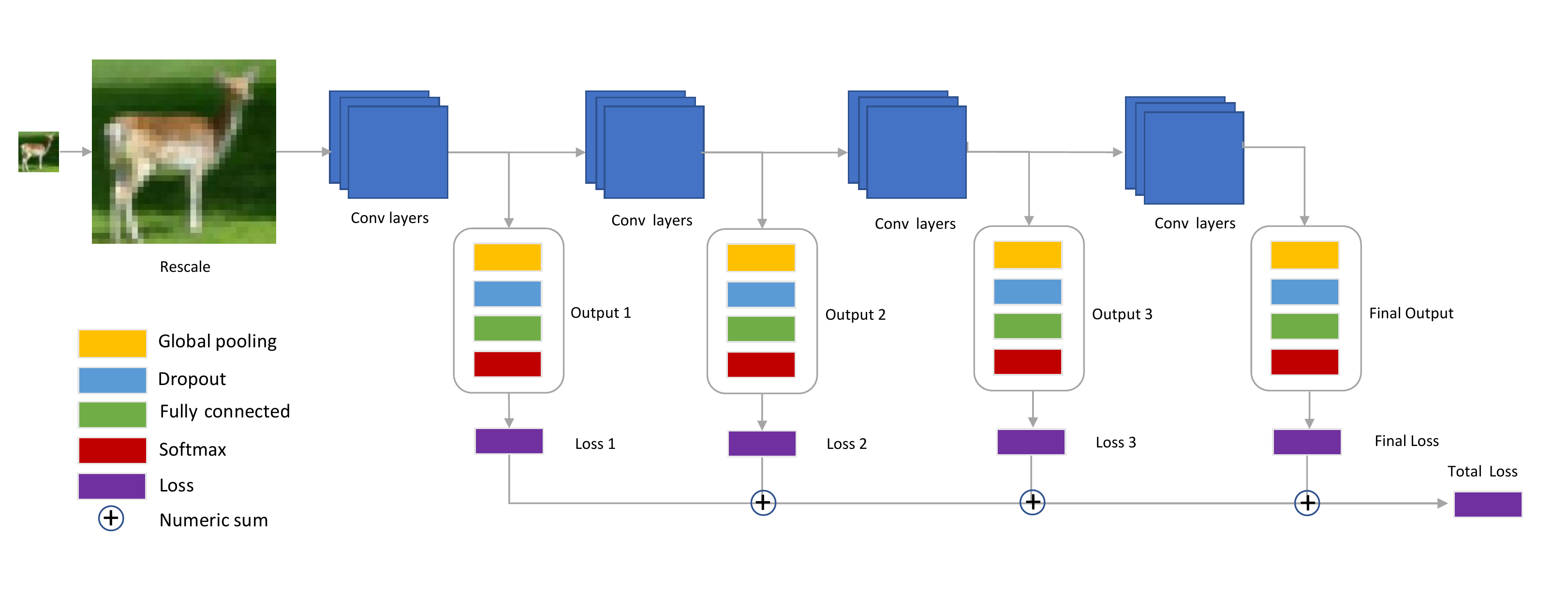}
 \caption{the Architecture of Elastic network}
 \label{fig:Architecture_JBnet}
 \end{center}
\end{figure*}

\section{Related Studies}

The proposed work is related to \textit{model regularization} (intermediate outputs and their respective losses add the constraint that also the features of the early layers should be useful for prediction), \textit{avoiding the gradient vanishing} as well as \textit{flexible swithing} between operating modes with different computational complexities. 

Regularization is a commonly used technique to avoid the over-fitting problem in DCNNs.  There are many existing regularization methods, such as, weight decay, early stopping, L1 and L2 penalization, batch normalization \cite{ioffe2015batch} and dropout \cite{srivastava2014dropout}. 
%In \cite{huang2016deep}, authors noticed it was possible to randomly drop some of the information paths during training, which selects stochastic neural network depth. 
Auxiliary outputs can also help with regularization and they have been applied in GoogLeNet \cite{szegedy2015going} with two auxiliary outputs. The loss function was the sum up of the two weighted auxiliary losses and the loss on the final layer. As a result, the auxiliary outputs increase the discrimination in lower layers and add extra regularization. In their later paper of Inception nets\cite{szegedy2016rethinking}, ImageNet-1K dataset was tested on both with and without auxiliary outputs Inception models, respectively. Experimental results showed that in the case with auxiliary outputs, classification accuracy was improved by 0.4\% in top-1 accuracy than the case without adapting intermediate outputs.

In addition to regularization, intermediate outputs are also expected to enhance the convergence during neural networks training. Deeply-supervised Nets (DSN) \cite{lee2015deeply} connected support vector machine (SVM) classifiers after the individual hidden layers. Authors declared that the intermediate classifiers promote more stable performance and improve convergence ability during training. 

More recently, many studies related to a more systematic approach of using intermediate outputs with neural networks have appeared. In Teerapittayanon \textit{et al.} \cite{teerapittayanon2016branchynet}, two extra intermediate outputs were inserted to AlexNet. This was shown to improve the accuracy on both the last and early outputs. The advantages of introducing auxiliary outputs has also been studied in Yong \textit{et al.} \cite{guo2016shallow}, where the authors compare different strategies for gradient propagation (for example, should the gradient from each output be propagated separately or jointly). Moreover, our recent work \cite{bai2018elastic} studied the same idea for regression problems mainly from the computational performance perspective. Here, we extend that work to classification problems and concentrate mainly on accuracy improvement instead of computational speed.

%In \cite{szegedy2016rethinking}, attaching outputs on intermediate layers can be also considered as an regularization method in DCNNs. In GoogleNet, the characteristics of the middle layer of the network are discriminative. the gradient vanishing problem can be mitigated by extracting the middle and lower features with using the auxiliary classifiers, and by adding auxiliary classifiers, it can also considered as regularization. Authors argue that the auxiliary classifiers act as regularizer. Gradient vanishing is one of the main challenges in deep neural network. For DCNNs like Inception which has 93 convolutional layer, has gradient vanishing problem for shallow layers when do backpropagation to update model weights. In ResNet, a skip connection is added to make backpropagation easier and mitigate the gradient vanishing problem. [should add explaining about applying intermediate outputs as acting gradient vanishing]

%MSDNet \cite{huang2017multi} uses DenseNet as the main model and add early-exits in the middle layers to achieve anytime prediction. For example, in MSDNet, 2 intermediate classifiers  are added after 2nd dense block, 3rd dense block respectively. Authors declare the introduction of intermediate classifiers harms the accuracy in ResNet main model and this issue can be mitigated when dense connectivity is used in DenseNet. The above observation isn't consistent with the other experiments in above mentioned  papers, and it also breaks our experiments result.
%what is the last sentence about???

%grammar

%separate gradient vanishing from regularization

\section{Proposed Architecture}
\begin{figure}
 \begin{center}
 \includegraphics[width=0.8\linewidth]{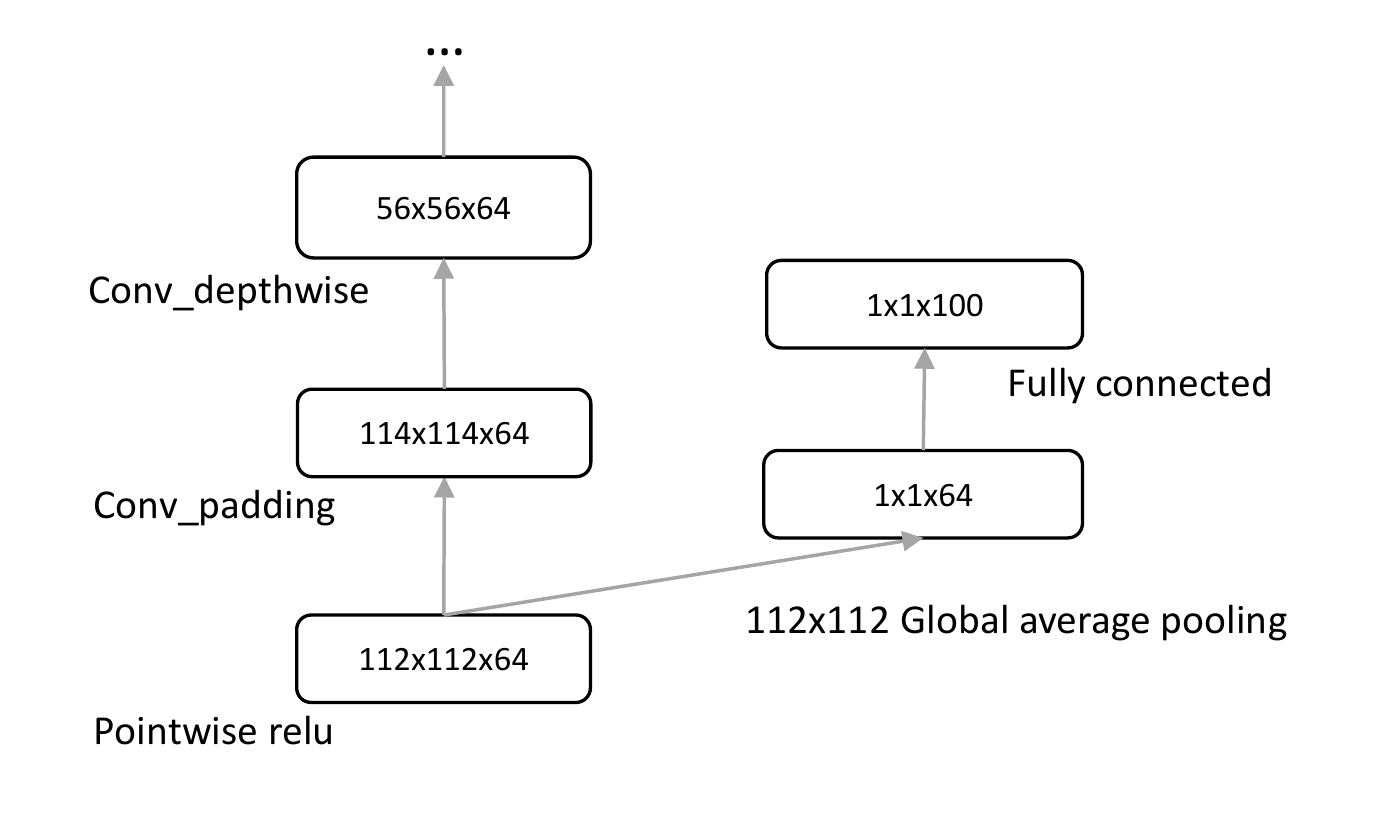}
 \caption{The intermediate output on the top of the first ``depth-conv'' block in MobileNet on CIFAR 100}
 \label{fig:intermediate-classifier}
 \end{center}
\end{figure}
%including constructing intermediate outputs, the number of outputs and corresponding positions on main models. %Then we explain the reason for mitigating gradient vanishing by using intermediate outputs.

%The JB Nets architecture is illustrated in Fig \ref{fig:Architecture_JBnet}. We attach outputs to intermediate layers in main models. For designing JB Nets, we address 3 main considerations, including, (1) construction of intermediate outputs, (2) the number of intermediate outputs and their positions, (3) loss function and model weights updating strategy.

\subsection{Attaching the intermediate outputs}
In the structure of the intermediate outputs shown in Fig. \ref{fig:Architecture_JBnet}, each unit of intermediate outputs consists of a global average pooling layer, a dropout layer, a fully connected layer and a classification layer with softmax function. %Dropout layers are recommended to use in \cite{szegedy2016rethinking}, authors argued that the dropout layers help regularization and avoid overfitting.
The insight behind the global average pooling layer is that generally the data size at the early layers is large (\textit{e.g.,} 224 $\times$ 224 $\times$ 3), and attaching a fully connected layer directly would introduce an unnecessarily large number of new parameters. %Here we set dropout rate equal to 0.2 during all of our training phrase. 
The same intermediate output structure is directly connected to the network after \textit{each} convolutional unit.

\begin{table*}[t]
\vspace{3mm}
	
	\caption{Testing error rate (\%) on CIFAR 10 and CIFAR 100.}
	%%\small
	%\addtolength{\tabcolsep}{-1pt}
	\begin{center}
    	\begin{threeparttable}
			\begin{tabular}{ l || c | c| c || c |c | c }
				\hline\hline
				 \multirow{2}{4em}{\textbf{Model} \textsuperscript{1}}& \multicolumn{3}{ c ||}{CIFAR-10} & \multicolumn{3}{ c }{CIFAR-100} \\ 
				\cline{2-7}
				& w/o Elastic Structure & w/ Elastic Structure & Improvement & w/o Elastic Structure & w/ Elastic Structure & Improvement\\
				\hline
                DenseNet-121    & 6.35 &  \textbf{5.44} &   14.3\%    & 24.48    &\textbf{23.44} & 4.3\%\\
                DenseNet-169    &8.14 &  \textbf{5.58}  &   31.5\%     & 23.06	& \textbf{21.65} & 6.1\%\\
                
                Inception V3    & 6.39  &  \textbf{4.48} &  29.9\%     & 24.98	&\textbf{21.83}  & 12.6\%\\
                MobileNet       & 10.37 & \textbf{7.82} &    24.6\%     & 38.60   & \textbf{25.22} & 34.7\% \\

                VGG-16	      &\textbf{7.71} &    8.20     &  -6.4\%     & 38.06 & \textbf{33.03}    & 13.2\% \\
%                 VGG-19    &\textbf{7.16} 	& 7.77 & \textbf{31.16} & 31.75 \\
                ResNet-50    &\textbf{5.54} 	& 6.19    &  -11.7\%     & \textbf{21.96}& 24.20       & -10.2\%  \\ 
                ResNet-50-5  	&	5.54	&	\textbf{5.39}	& 2.7\% 	&	21.96	&  \textbf{21.54} &  1.9\%   \\
               \hline
                PyramdNet + ShakeDrop\textsuperscript{2} 	&   2.3   &  -   & - &   12.19    &  - & - \\

			\end{tabular}
            \begin{tablenotes}\footnotesize
% 				\item[1] The $\alpha$ parameter in MobileNet increases or decreases the number of filters in each layer.  
                \item[1]  All the models we use are pretraiend on ImageNet
                
                \item[2] the state of the art accuracy \cite{yamada2018shakedrop}
			\end{tablenotes}
		\end{threeparttable}
	\end{center}\vspace{-0.5cm}%\vspace{-0.35cm}%\vspace{-0.45cm} 
    \label{tbl:equipos}
\end{table*}

\subsection{The number of outputs and their positions}
% * <yipersevere@outlook.com> 2018-09-18T22:40:08.428Z:
% 
% 也许我需要修改 the number of outputs and their positions 这个子标题，换成一个比如说：构建JBNet 实例？
% 
% ^ <yipersevere@outlook.com> 2018-09-18T22:41:14.739Z.
%The performance of JB Net is directly influenced by the number and the position of intermediate outputs. 
Since network learns low level features on the early layers, the total number of weights allocated to the intermediate outputs is crucial to the final result. The number and the position of intermediate outputs in Elastic Nets are separately designed based on the different original network structures (e.g. network depth, network width).
%shown in Table \ref{tbl:classifierslocation}. 
%ResNet repeatly uses ResBlock to construct different depth of ResNet, like ResNet-50, ResNet-101. %

Although our general principle is to add auxiliary outputs after \textit{every} layer, the particular design of the widely used networks require some adjustment. The network specific adjustments for each network in our experimentation are described next. 

\textbf{Elastic-Inception V3} is based on the Inception-V3 architecture \cite{szegedy2016rethinking} having  94 convolutional layers arranged in 11 Inception blocks. We add intermediate outputs after the ``concatenate'' operation in each Inception block except the last one. Together with the final layer, in total, there are 11 outputs in the resulting Elastic-Inception V3 model.

\textbf{Elastic-DenseNet-121} is based on the DenseNet-121 architecture \cite{huang2017densely} with the  \textit{growth\_rate} hyperparameter equal to 32. DenseNet-121 consists of 121 convolutional layers grouped into 4 dense blocks and 3 transition blocks. We attach each intermediate output after the ``average pooling'' layer of  each transition block. We apply the same strategy for building Elastic-DenseNet-169, where the backbone is similar but deeper than the DenseNet-121. Totally, there are 4 outputs in both Elastic-DenseNet-121 and Elastic-DenseNet-169.
%how much in total respectively?

\textbf{Elastic-MobileNet} is based on the MobileNet architecture \cite{howard2017mobilenets} with hyperparameters $\alpha=1$ and $\rho=1$. Here, an intermediate output is added after the ``relu'' activation function in each depthwise convolutional block. MobileNet has 28 convolutional layers and consists of 13 depthwise convolutional blocks. Elastic-MobileNet has 12 intermediate outputs besides 1 final classifier. We illustrate a part of the intermediate output structure from Elastic-MobileNet in Fig \ref{fig:intermediate-classifier}.

\textbf{Elastic-VGG-16} design is based on VGG16 architecture \cite{simonyan2014very}. Intermediate outputs are attached after each of the ``maxpooling'' layers. In total, there are 5 outputs in Elastic-VGG-16.% The same strategy is applied for VGG-19, 5 outputs are in JB-VGG-19.

\textbf{Elastic-ResNet-50} is designed based on ResNet50 \cite{he2016deep}. Intermediate outputs are attached after each of the ``add'' operation layers in the \textit{identity\_block} and the \textit{conv\_block}. The \textit{conv\_block} is a block that has a convolutional layer at shortcut, \textit{identity\_block} is a block with no convolutional layer at shortcut. In total, there are 16 total outputs in Elastic-ResNet-50.

% \begin{table}[htbp]
% \renewcommand{\arraystretch}{1}
% \caption{location of intermediate outputs on different neural networks}
% \label{tbl:classifierslocation}
% \centering
% \begin{tabular}{c|c|c|c}
% \hline
% \textbf{Model} & \multicolumn{1}{ c |}{\textbf{No.Layers }}& \multicolumn{1}{ c |}{\textbf{No.Classifiers}}& \multicolumn{1}{ c }{\textbf{Positions}}  \\ 
% \hline\hline
% JB-VGG 16          & 13   & 5    & {4, 7, 11, 15, 18}\\
% %JB-Squeezenet      & 26   & 14   & {after each fire module} \\
% JB-Mobilenet       & 27   & 13   & {}\\
% %JB-ResNet-50       & 50   & 17   & \\
% JB-Inception V3    & 93   & 12   & {19, 20, ...30} \\
% JB-DenseNet 121    & 121  & 4    & {after dense block}\\
% %JB-DenseNet 169    & 169  & 4    & {after each dense block}\\
% \hline
% \end{tabular}
% \end{table}
% * <yipersevere@outlook.com> 2018-09-19T10:55:19.550Z:
% 
% 也许我可以不要　table1, 也就是说删除掉关于不同网络的intermediate outputs location and numbers, 可以直接在段落中描述这个
% 
% ^ <yipersevere@outlook.com> 2018-09-19T10:56:18.285Z.

% * <yipersevere@outlook.com> 2018-09-20T14:51:30.364Z:
% 
% I will change the picture soon. 
% 
% ^ <yipersevere@outlook.com> 2018-09-20T14:51:49.285Z.

\subsection{Loss function and weight updates}

Forward and backpropagation are executed in a loop  during the training process. In the forward step, denote the predicted result at intermediate output $i\in \{1,2,\ldots, N\}$ by $\hat{\bf y}^{(i)} = [\hat{y}_1^{(i)}, \hat{y}_2^{(i)}, \ldots, \hat{y}_C^{(i)}]$, where $C$ is the number of classes. We assume that the output yields from the softmax function:
\begin{align}\label{Eq:softmax}
\hat{y}_k^{(i)} &= \text{softmax}({\bf z}) 
= \frac{\exp\left(z_i^{(k)}\right)}{\displaystyle \sum_{c\in \{1,\ldots, C\}} \exp\left(z_i^{(c)}\right)}, 
\end{align}
where ${\bf z}$ is a vector output of the last layer before the softmax.

As shown in Fig. \ref{fig:Architecture_JBnet}, loss functions are connected after softmax function. Denote the negative log-loss at $i$'th  intermediate output by
\begin{align}\label{Eq:single-loss}
L_{i}(\hat{\bf y}^{(i)}, {\bf y}^{(i)}) = 
&-\frac{1}{C} \displaystyle \sum_{c\in \{1,\ldots, C\}} y_c^{(i)} \log \hat{y}_c^{(i)},
\end{align}
where ${\bf y}^{(i)}$ is the one hot ground truth label vector at output $i$, and $\hat{y}_c^{(i)}$ is the predicted label vector on the $i$'th output. 
The final loss function $L_{total}$ is the weighted sum of losses at the intermediate outputs:
\begin{equation}\label{Eq:total-loss}
% L\left (y_{exit_{n}}, y\right. ;\theta )=\sum_{n=1}^{N}w_{n}L \left ( y_{exit_{n}}, y\right.;\theta ).
L_{total} = \sum_{i=1}^{N}w_{i}L_{i}\left(\hat{\bf y}^{(i)}, {\bf y}^{(i)}\right),
\end{equation}
with weights $w_i$ adjusting the relative importance of individual outputs.
In our experiments, we set $w_{i}=1$ for all $i \in \{1,\ldots,N\}$.

\section{Experiments}
%In this section, we begin by describing dataset and experiment setting. Then we introduce our experiment results and.
%\subsection{Datasets and Experiment setting}\label{AA}

We test our framework on the commonly used CIFAR10 and CIFAR100 datasets \cite{krizhevsky2009learning}. Both of them are composed of 32$\times$32 color images. CIFAR10 has 10 exclusive classes, and there are 50,000 and 10,000 images in training and testing sets, respectively. CIFAR100 contains 100 classes, and each class has 500 training samples and 100 testing samples. In our experiments, the original training set is split into two sets with 80\% as our training set and  20\% as our validation set. Our training set has 40,000 images and validation set contains 10,000 images. Our test set still uses the original CIFAR test set, which has 10,000 images.
% what kind of classes are there, where is the data gather from?

% \textbf{ImageNet}
% The dataset includes images of 1000 classes.

% \textbf{Tiny ImageNet.}
% Tiny ImageNet dataset consists of 200 classes of objects, the training set contains 100,000 images, and testing set contains 10,000 images, each class has 500 training images and 50 testing images. All images are 64$\times$64 pixels.

\subsection{Experimental setup and training}

At network input, the images are resized to 224$\times$224$\times$3 and  we normalize the pixels to the range [0,1]. Data augmentation is not used in our experiment. 
We construct Elastic Nets using Keras platform \cite{chollet2015keras}. The weights for different Elastic Nets are initialized according to their original networks pretrained on ImageNet, except the output layers that are initialized at random.

Since we introduce a significant number of random layers at the network outputs, there is a possibility that they will distort the gradients leading into unstable training. Therefore, we first train \textit{only }the new random layers with the rest of the network frozen to the imagenet-pretrained weights. This is done for 10 epochs with a constant learning rate $10^{-3}$.

After stabilizing the random layers, we unfreeze and train all weights with an initial learning rate $10^{-3}$ for 100 epochs. Learning rate is divided by 10 after the decaying loss on validation set staying on plateau for 10 epochs. All models are trained using SGD with mini-batch size 16 and a momentum parameter 0.9. The corresponding original networks are trained with the same settings.
Moreover, data augmentation is omitted, and the remaining hyperparameters are chosen according to the library defaults \cite{chollet2015keras}.

\begin{table}[htbp]
  \centering
  \caption{Testing error rate (\%) of Elastic-DenseNet-169 with 4 different depth models, DenseNet-169 and DenseNet-121 on CIFAR 100.}
    \begin{tabular}{l|c|c|c}
    \hline
    \textbf{Model}& \textbf{\# conv layer}&{\textbf{params}}&{\textbf{error}}\\ 
    \hline
    Elastic-DenseNet-169-output-14      &        14          &       0.39M &   73.37      \\
    Elastic-DenseNet-169-output-39      &        39          &       1.47M &   48.96      \\
    Elastic-DenseNet-169-output-104      &        104         &       6.71M &   22.86      \\
    Elastic-DenseNet-169-output-168      &        168         &       20.90M &   21.65      \\
     DenseNet-169                  &       168           &      20.80M  & 23.06\\
    DenseNet-121                   &        120         &     12.06M   &   24.48      \\
%     JB-DenseNet-201-3      &        136           &      --M &   --      \\
%     JB-DenseNet-201-4      &        200         &       --M &   --      \\
    \hline
    \end{tabular}
  \label{tbl:DenseNetDifferentDepthModel}
\end{table}

\subsection{Results}\label{BB}

In our experiment, we wish to compare the effect of intermediate outputs with their corresponding original nets. 
In our framework, each intermediate output can be used for prediction. For convenience, we only compare the last layer output as it is likely to be the most accurate prediction. The results on CIFAR test datasets are shown in Table \ref{tbl:equipos}.

From Table \ref{tbl:equipos} we can see that in general  the Elastic structured networks perform better compared to most of their back-bone architectures for both datasets. More specifically, the use of intermediate outputs and their respective losses improves the accuracy for all networks except VGG-16 and ResNet-50 for both datasets. Moreover, for CIFAR-100,  ResNet-50 is the only network that does not gain in accuracy from the intermediate outputs. 

In CIFAR100, Elastic-DenseNet-169 achieves the lowest error rate (5.58\%) among all models, which is 6.1\% lower error rate than DenseNet-169. Elastic-DenseNet-121 and Elastic-Inception V3 have the better result than the original models as well, which reduce the error rate by 4.3\% and 12.6\% respectively. In CIFAR 10, Elastic-DenseNet-169 decreases 31.5\%  error compared to DenseNet-169. Elastic-DenseNet-121 and Elastic-Inception V3 outperform the back-bone networks by 14.3\% and 29.9\%.

%  on CIFAR 10 and 1.04\% lower error rate on CIFAR 100. JB-DenseNet-169 achieves the lowest error among all models both on CIFAR 10 (5.58\%) and CIFAR 100 (21.65\%) in our experiments, which reduces 2.56\% and 1.41\% error rate than DenseNet-169 on CIFAR 10 and CIFAR 100 respectively. For JB-Inception V3, JB-Inception V3 has 1.91\% lower error than Inception V3 in CIFAR 10 and 2.64\% lower error in CIFAR 100. 

In relative terms, the highest gain in accuracy is obtained with MobileNet. 
Namely, Elastic-MobileNet achieves testing error of 25.22\%, exceeding the backbone MobileNet (38.60\%) by 35\% relative improvement in CIFAR 100, and it's the biggest performance improvement compared to all other Elastic Nets. In CIFAR 10, Elastic-MobileNet also reduces the error by 25\%.

There are two exceptions to the general trend: VGG16 and ResNet50 are not always improving in accuracy with the intermediate outputs. One possible reason is that VGG is shallow convolutional neural network. There is less gradient vanishing issue (over-parameters). Therefore, adding intermediate outputs effect positively less. 
The same reason probably holds for the ResNet-50 architecture, as well. Although the network itself is deep, there are \textit{shortcut connections} every few layers, which help to propagate the gradient more effectively. Due to these reasons, the added layers do not have equally significant improvement as with the other networks.

%JB-ResNet-50 with 17 outputs performs less good with 0.65\% and 2.24\% higher error than ResNet-50 on CIFAR 10 and CIFAR 100 respectively. 

Additionally, the number of intermediate outputs is clearly the highest with the ResNet-50. In order to verify whether this might be a part of the reasons why ResNet-50 does not gain from our methodology, we decreased their number from 17 to 4 by removing all but the 2nd, 6th, 9th, and 12th intermediate outputs. This structure is denoted as "Elastic-ResNet-50-5" in Table \ref{tbl:equipos}. In this case, Elastic-ResNet-50-5 outpeforms the original ResNet-50 slightly by decreasing error by 2.7\% and 1.9\% on CIFAR 10 and CIFAR 100, respecively. 

%For example, in CAFAR 10, JB-VGG-16 increases error rate by 0.41\% that VGG-16 model, JB-VGG-19 suffers 0.61 higher error rate than VGG-19. In CIFAR 100, JB-VGG-16 and JB-VGG-19 have worse results than the originals, with 0.56\% and 0.59\% higher error rate respectively. 

So far, we have concentrated only on the accuracy of the last layer of the network. However, the accuracy of the intermediate layers is also an interesting question. In our earlier work \cite{bai2018elastic} we have discovered that also the other late layers are relatively accurate in regression tasks. Let us now study this topic with one of the experimented networks.

We study the case of DenseNet-169, because it has fewer intermediate outputs than the others and is thus easier to understand. The Elastic extension of this network has three intermediate outputs (after layers 14, 39 and 104) and the final output (after layer 168). We compare the Elastic extension of DenseNet-169 against the normal DenseNet-169 in Table \ref{tbl:DenseNetDifferentDepthModel}. The errors of all four outputs for test data on CIFAR 100 are shown on the rightmost columns and show that the best accuracy indeed occurs when predict from the ultimate layer. On the other hand, also the penultimate layer is more accurate than the vanilla DenseNet, and has remarkably fewer parameters. Moreover, using the 104'th layer output of the DenseNet turns out to be also more accurate than DenseNet-121, which is roughly the same depth but with significantly more parameters. 

%%%%%%%%%%%%%%%%%%%%%%%%%%%%%%%%%%%%%%%%%%%%%%%%%%%%%%%%%%%%

\iftrue

% \begin{figure*}[t]
% 	\centering
%     {%
%     \subfloat[]{{\includegraphics[width=0.35\linewidth]{figures/Train_Test_InceptionV3_CIFAR_100_00.pdf} }}%
%     \qquad
%     \subfloat[]{{\includegraphics[width=0.35\linewidth]{figures/Train_Test_DenseNet121_CIFAR_100_00.pdf} }}
%     }
%     \caption{Training and testing error evolution of Inception V3 and DenseNet-121 with and without Elastic structure on CIFAR 100. Dotted curves denote training error, and solid curves denote testing error. (a): Inception V3 and Elastic-Inception V3; (b): DenseNet-121 and Elastic-DenseNet-121 }%
%     \label{fig:testErrorEpochs}
% \end{figure*}

% \begin{figure}[htbp]
%  \begin{center}
%  \includegraphics[width=0.7\linewidth]{figures/CIFAR_100_accuracy_JB-DenseNet-169_DenseNet-121.pdf}
%  \caption{Testing error of Elastic-DenseNet-169 with generating 4 different depths models and DenseNet-121 on CIFAR 100}
%  \label{fig:JBDenseNet169testingErrorCIFAR100}
%  \end{center}
% \end{figure}

\begin{figure}[htbp]
 \begin{center}
 \includegraphics[width=1\linewidth]{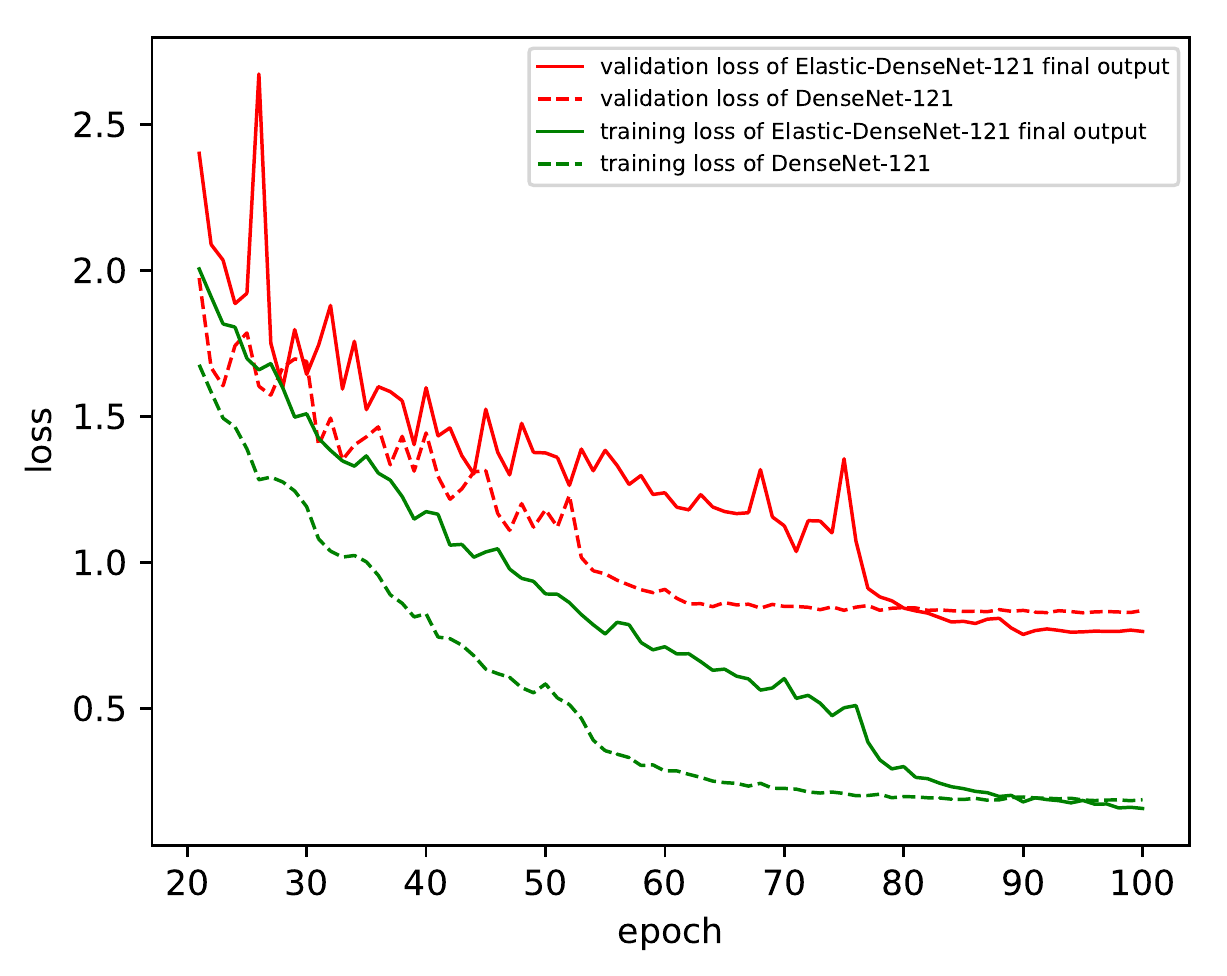}
 \caption{Comparison of training and validation loss curves of Elastic-DenseNet-121 final output and DenseNet-121 on CIFAR 10 dataset}
 \label{fig:DenseNet121LossForCIFAR10}
 \end{center}
\end{figure}

% \subsection{Loss curves}

It is of interest to see how the learning behavior changes when the intermediate outputs are added. To this aim, we plot the training and validation loss curves of Elastic-DenseNet-121 and DenseNet-121 on CIFAR10 dataset, as shown in  Fig. \ref{fig:DenseNet121LossForCIFAR10}. In the figure, we can see that the  Elastic-DenseNet-121 model has lower training loss value than DenseNet-121 without the intermediate outputs, as seen in the two green curves. For the validation set, Elastic-DenseNet-121 also achieves a lower loss value than DenseNet-121, as shown in the two red curves. This observation is coherent with Table \ref{tbl:equipos} where we can see that Elastic-DenseNet-121 has also a higher test accuracy than DenseNet-121 on both CIFAR10 and CIFAR100 test sets. Moreover, the learning curves indicate that the elastic structure learns to classify \textit{slower} but leads into a better final accuracy than the plain DenseNet-121 without the elastic structure. This also leads us to speculate whether the effect of the intermediate outputs, in fact, to inject noise to the network---similarly as dropout---and is a subject of further study in this domain.

\fi

\section{Discussion and conclusion}

This paper investigates employing intermediate outputs while training deep convolutional networks. When neural networks become to deeper, the gradient vanishing and over-fitting problems result in decreasing  classification accuracy. To mitigate these issues, we proposed to feed the gradient directly to the lower layers. In the experimental section we showed that this yields significant accuracy improvement for many networks. There may be several explanations for this behavior, but avoiding the vanishing gradient seems most plausible, since the residual networks with shortcut connections do not gain from the intermediate outputs. 

Interestingly, we also discovered, that using early exits from deep networks can be more accurate than the final exit of a network of equivalent depth. In particular, we demonstrated that using the 104'th exit of the DenseNet-169 becomes more accurate than the full DenseNet-121 when trained with the proposed framework.

% However, Joint Branches Networks architecture is general and can be used for other neural network and other tasks like object detection. 
% We consider adding intermediate classifiers can mitigate gradient vanishing problem and attaching classifiers to intermediate classifiers can be considered as one method of regularization.

%@inproceedings{bai2018elastic,
%author={masked for double blind review},
%title = {masked},
%booktitle = {masked},
%year = {2018}
%}

{
\bibliographystyle{IEEEtran}
\bibliography{IEEEabrv,AICAS_conference.bib}
}
% insert extra pictures
% \begin{figure}[htbp]
% 	\centering
%     {%
%     \subfloat[]{{\includegraphics[width=0.8\linewidth]{figures/testing_SqueezeNet-MobileNet-alpha-0_75_CIFAR_100_00.png} }}}%  
%     \caption{Performance on different Elastic Framework based networks. Lower is better. }%
% \end{figure}

\end{document}